\documentclass{bmvc2k}

\usepackage{graphicx}
\usepackage{amsmath}
\usepackage{amsthm}
\usepackage{booktabs}
\usepackage{algorithm}
\usepackage{algorithmic}
\usepackage{multirow}
\usepackage{bbding}
\usepackage{amssymb}
\usepackage{wrapfig}
\usepackage{ulem}
\usepackage{color}

\title{Highly Efficient Natural Image Matting \footnote{ \textbf{Bo Li is equal contribution to the first author. Corresponding authors are Bo Li and Lv Tang.}} }

\addauthor{Yijie Zhong}{dun.haski@gmail.com}{3}
\addauthor{Bo Li}{libraboli@tencent.com}{1}
\addauthor{Lv Tang}{luckybird1994@gmail.com}{3}
\addauthor{Hao Tang}{hao.tang@vision.ee.ethz.ch}{2}
\addauthor{Shouhong Ding}{ericshding@tencent.com}{1}

\addinstitution{
 Youtu Lab, \\
 Tencent, \\
 Shanghai, China
}
\addinstitution{
Computer Vision Lab, \\
ETH Zurich, \\
Switzerland
}
\addinstitution{
Independent Researcher
}

\runninghead{Zhong et al.}{Highly Efficient Natural Image Matting}

\begin{document}

\maketitle

\begin{abstract}
Over the last few years, deep learning based approaches have achieved outstanding improvements in natural image matting. However, there are still two drawbacks that impede the widespread application of image matting: the reliance on user-provided trimaps and the heavy model sizes. In this paper, we propose a trimap-free natural image matting method with a lightweight model. With a lightweight basic convolution block, we build a two-stages framework: Segmentation Network (SN) is designed to capture sufﬁcient semantics and classify the pixels into unknown, foreground and background regions; Matting Refine Network (MRN) aims at capturing detailed texture information and regressing accurate alpha values. With the proposed cross-level fusion Module (CFM), SN can efficiently utilize multi-scale features with less computational cost.  Efficient non-local attention module (ENA) in MRN can efficiently model the relevance between different pixels and help regress high-quality alpha values. Utilizing these techniques, we construct an extremely light-weighted model,  which achieves comparable performance with ~1\% parameters (344k) of large models on popular natural image matting benchmarks. 
\end{abstract}

\section{Introduction}
The natural image matting is an important task in computer vision, which serves as a prerequisite for a broad set of applications, including image or video editing, compositing and film post-production~\cite{DBLP:conf/cvpr/TangAOGA19,DBLP:conf/cvpr/SenguptaJCSK20,DBLP:journals/corr/abs-2012-07810,tang2020xinggan,tang2020local,tang2019multi,tang2021attentiongan,tang2020dual,tang2020bipartite}. Following~\cite{DBLP:conf/siggraph/PorterD84}, the natural image $I$ is defined as a convex combination of foreground image $F$ and background image $B$ at each pixel $i$ as:
\begin{equation}
I_i = \alpha_iF_i + (1-\alpha_i)B_i, \alpha_i \in [0,1],
\end{equation}
where $\alpha_i$ is the alpha value at pixel $i$ that denotes the opacity of the foreground object. Mathematically, image matting requires expressing pixel colors in the transition regions 
\begin{wrapfigure}{r}{0.5\linewidth}
    \centering
    \includegraphics[width=\linewidth]{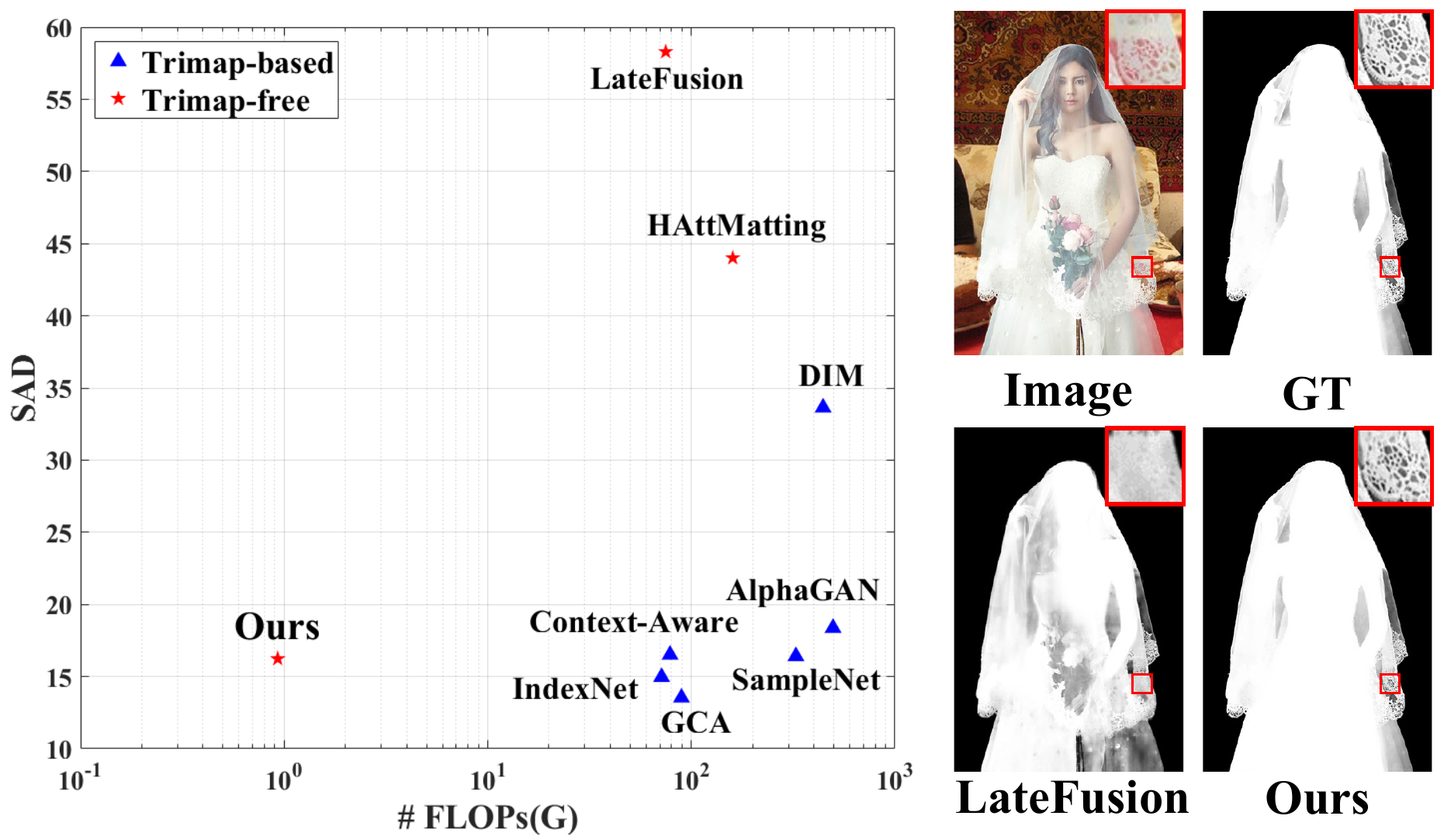}
    \caption{Comparison with SOTA methods. Best viewed by zooming in. Smaller SAD and FLOPs mean better performance and less computational cost. }
    \label{fig:intro}
\end{wrapfigure}
from foreground to background as a convex combination of their underlying foreground and background colors. Since neither the foreground and background colors nor the opacities are known, estimating the opacity values is a highly ill-posed problem. To alleviate the difficulty of this problem, typically a trimap is provided in addition to the original image~\cite{DBLP:conf/cvpr/AksoyAP17}. The trimap can be provided by user or generated from the network. The trimap is a rough segmentation of the input image into foreground($\alpha_i=1$), background($\alpha_i=0$), and regions with unknown opacity($\alpha_i\in(0,1)$) which like other segmentation tasks~\cite{DBLP:conf/iccv/LiSLWH19,DBLP:conf/aaai/LiSG19,DBLP:conf/ijcai/0061STSS19,DBLP:conf/mm/LiSWL19,DBLP:journals/mta/LiSXWY20,DBLP:conf/accv/Tang020,DBLP:conf/pcm/LiSHX17,Tang_2021_ICCV}. From then on, many traditional methods~\cite{DBLP:conf/cvpr/ChuangCSS01,DBLP:journals/tog/SunJTS04,DBLP:conf/cvpr/LevinLW06,DBLP:conf/cvpr/WangC07,DBLP:conf/cvpr/HeRRTS11,DBLP:conf/cvpr/ShahrianRPC13,DBLP:conf/cvpr/AksoyAP17} are proposed for matting task. These methods leverage color features to predict alpha mattes, which will produce obvious artifacts when foreground and background share similar colors.

Deep learning has recently demonstrated success in many computer vision applications. Recent researches use deep neural networks to improve the performance of natural image matting. Compared to traditional methods, deep-learning matting methods~\cite{DBLP:conf/eccv/ChoTK16,DBLP:conf/cvpr/XuPCH17,DBLP:conf/cvpr/TangAOGA19,DBLP:conf/iccv/CaiZFHLLLWS19,DBLP:conf/iccv/Hou019,DBLP:conf/iccv/0003DSX19,DBLP:conf/aaai/LiL20a} have achieved remarkable performance and further advance the matting field.  However, the widespread application of image matting still faces two major challenges: first, most methods rely on user-provided trimaps as assistant input, which is not easy to use for novice users without expertise in digital matting; second, their heavy model sizes make them resource-hungry and hardly applicable on low-power devices with limited storage/computational capability. 

To address first problem,~\cite{DBLP:conf/cvpr/ZhangGFRHBX19,DBLP:conf/cvpr/QiaoLYZXZW20} design trimap-free natural image matting models.~\cite{DBLP:conf/cvpr/ZhangGFRHBX19} investigated semantic segmentation variant for foreground and background weight map fusion to obtain alpha values.~\cite{DBLP:conf/cvpr/QiaoLYZXZW20} explores advanced semantics and appearance cues synthetically to achieve high-quality alpha values. However, to capture sufﬁcient semantics, these two methods utilizate large backbones (DenseNet-201 and ResNeXt101), which make it difficult to apply these methods to some devices with limited storage/computational capability. To address second problem, ~\cite{DBLP:journals/spl/YoonPC20} designs a light-weighted natural image matting model with the help of knowledge distillation. However, this method also needs extra trimaps, which is not user-friendliness. Hence, designing a trimap-free natural image matting method with a light-weighted model is barely explored. 

Without the user-provided trimap, there are two basic tasks for automatic natural image matting: identify the foreground, background and unknown regions in the image; predict the precise alpha values of unknown regions. The first one is a classic classiﬁcation task while the second task is typical regression task. As discussed in ~\cite{DBLP:conf/cvpr/ZhangGFRHBX19}, it is hard to handle the two tasks simultaneously with a single structure, we thus propose a two-stage end-to-end deep learning framework for trimap-free light-weighted matting task. The first stage is designed to capture sufﬁcient semantics to identify the foreground, background, and unknown regions. Then under the guidance of the first stage, the second stage can give the precise alpha values of unknown regions. Although such two-stage network design has appeared before (e.g. \cite{DBLP:conf/mm/ZhuCWLZT17}, \cite{ DBLP:conf/crv/HuC19}, \cite{DBLP:conf/mm/ChenGXZYG18}), it is an intutitive idea. The most important is that \cite{DBLP:conf/mm/ZhuCWLZT17} designs redundant structures besides low-resolution input ($128\times 128$), and the strong portrait prior limit their application prospects. Then the core problem lies in how to accomplish these two tasks with light-weighted network structures.

We approach this core problem from two perspectives: a light-weighted backbone and efficient feature aggregation/enhancement strategies of each stage. To be specific, we first build a light-weighted basic block with some popular efficient convolutional operations. Then we stack these basic blocks to construct light-weighted U-Net like backbones for two stages. The first stage is segmentation network (SN), and the second stage is matting refine network (MRN). In the first stage, we design a cross-level fusion module (CFM) to directly fuse the high-level semantic feature with low-level feature, which can improve the discriminability of feature representation. In the second stage, we design a novel efficient non-local attention module (ENA) to model the relevance between different pixels in encoder/decoder feature and image feature extracted from octave block (OCBlock), which can help regress high-quality alpha values. Moreover, in matting task, modeling the relevance of different pixels in long-range is not as important as modeling the relevance of pixels in short-range. So the proposed ENA first calculates long-range relevance with sampling strategy, then directly calculates short-range relevance. This design meets the principle of a light-weighted model. As can be seen in Fig.\ref{fig:intro}, compared with LateFusion~\cite{DBLP:conf/cvpr/ZhangGFRHBX19}, our proposed method can predict a better result with accurate semantics and alpha value.

Our major contributions can be summarized as:
\begin{itemize}

\item We make one of the earliest effort on designing a light-weighted trimap-free natural image matting method, which expands matting to much wider application scenarios.

\item We construct two U-Net like backbone with light-weighted basic blocks for the proposed two-stage framework and propose two corresponding efficient feature aggregation/enhancement strategies CFM and ENA for two stages to better identify the unknown regions and give precise alpha values.

\item We perform extensive experiments to demonstrate the proposed method refreshes the SOTA performance on Composition-1k and Distinctions-646 testing sets with ~1\% parameters (344k) of SOTA large models.
\end{itemize}

\begin{figure}[t]
    \centering
    \includegraphics[width=\textwidth]{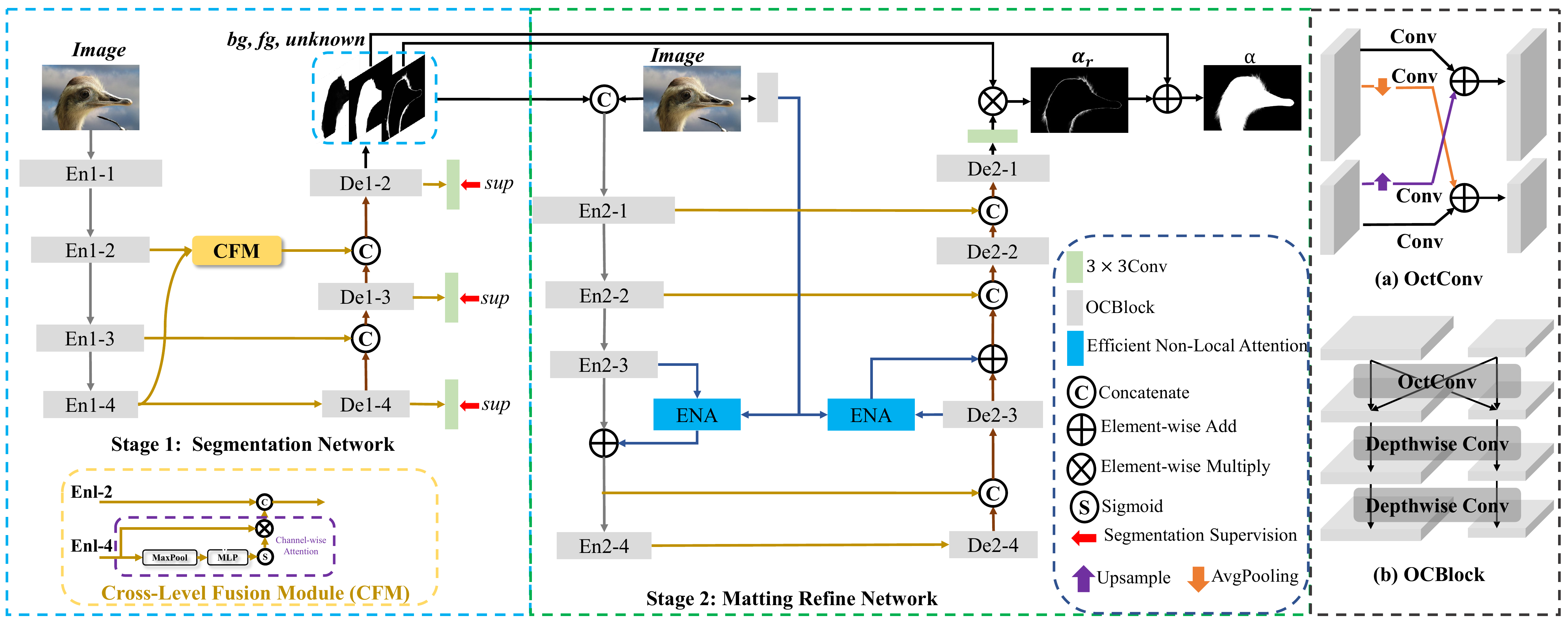}
    \caption{The framework of the our trimap-free light-weighted natural image matting model.}
    \label{fig:pipeline}
\end{figure}

\section{Method}
In this section, we ﬁrst describe the overall architecture of the proposed trimap-free light-weighted natural image matting network, then introduce our proposed light-weighted backbone, SN and MRN. The architecture of the proposed approach is illustrated in Fig.\ref{fig:pipeline}.

\subsection{Network Overview}
SN has a simple U-Net like Encoder-Decoder architecture, and the backbone is formed by stacking our proposed OCBlock. Similar to ResNet-50 \cite{DBLP:conf/cvpr/HeZRS16}, the proposed backbone has four levels, named as En1-1, En1-2, En1-3, En1-4. The number of OCBlock in each level is 3, 4, 6, 4. Because En1-1 is too close to the input and its resolution is too high, to save computational cost, we only use the last three levels features for the following process. Cross-level fusion module (CFM) is added between En1-4 and En1-2 to improve the discriminability of feature representations. SN has 3 decoders, and each decoder is stacked by 1 OCBlock. Decoder fuses the processed features from the encoders and the upsampled features from the previous decoder level in a bottom-up manner. The output of each decoder is defined as De1-4, De1-3, and De1-2. Finally, the network generates the 3-channel classification logits $T$ by a $3\times3$ convolution.

MRN has a basic Encoder-Decoder architecture, and the input of MRN is predicted logits $T$ and an image. The backbone is also stacked by our proposed OCBlock, and all encoder levels contain 2 OCBlocks, respectively. To accurately regress alpha values in unknown regions, we stick two efficient non-local attention (ENA) module to the encoder and decoder symmetrically in our stem. Moreover, image features extracted from OCBlocks contain more detailed texture information, which can help regress alpha values. Thus, we utilize the image features extracted from OCBlocks to guide the information flow on Encoder/Decoder features. MRN has 4 decoders, and each decoder is stacked by 1 OCBlock. Decoder fuses the features in each encoder and the upsampled features from the previous level in a bottom-up manner. Finally, together with the unknown and foreground maps predicted by SN, MRN can regress a 1-channel result $\alpha$.

\subsection{Light-weighted backbone}
Recently, some researchers have designed various light-weighted backbones~\cite{DBLP:journals/corr/HowardZCKWWAA17,DBLP:conf/cvpr/Chollet17,DBLP:conf/cvpr/ZhangZLS18,DBLP:conf/icassp/LiSTH19,DBLP:conf/icassp/TangLWXD21} with special convolution operations. A classic example is MobileNet which constructs a light-weighted backbone by using the depthwise convolution. However, these light-weighted models are designed for classification, it is not appropriate to directly use them for matting task. 
Unlike classification task, matting needs multi-scale information to capture sufﬁcient semantics for pixel-level prediction. Some researchers try to perform extra multi-scale feature extraction after backbone, like atrous spatial pyramid pooling (ASPP), global convolutional network (GCN)~\cite{DBLP:conf/cvpr/PengZYLS17}, which need more computational cost. Inspire by MobileNet, we want to use depthwise convolution to build our lightweight backbone. But only using depthwise convolution cannot capture sufficient multi-scale information. To address this problem, by leveraging the OctConv \cite{DBLP:conf/iccv/Chen0XYKRYF19}, we design our 
lightweight OCBlock, and stack OCBlocks to construct a light-weighted backbone. To be specific, we replace a depthwise convolution with a vanilla OctConv. The vanilla OctConv \cite{DBLP:conf/iccv/Chen0XYKRYF19} is presented to conduct the convolution operation on features across high/low scales, as shown in Fig.\ref{fig:pipeline}(a). In OctConv, features from different scales are fused to generate output. The proposed OCBlock is shown in Fig.\ref{fig:pipeline}(b), which combines the advantages of depthwise convolutions and OctConv to achieve a better weight/performance balance. 

\vspace{-0.3cm}
\subsection{Segmentation Network}
SN plays the role of capturing sufﬁcient semantics and classifying the pixels into unknown, foreground, and background regions with light-weighted backbone. Thus, learning discriminant feature representations with less computational cost is essential. To achieve this goal,  we first utilizate our proposed light-weighted backbone to obtain multi-scale feature representations. Moreover, we utilize CFM module to leverage the advantages of features at high-level to learn more discriminant feature representations. \\
\textbf{Cross-Level Fusion Module.} High-level feature from En1-4 contains much semantic information, which can make the network better distinguish the appearance difference between unknown, foreground, and background regions. However, our proposed SN is a U-like architecture, when information is progressively returned from the high-level to the low-level, the high-level semantic information is gradually diluted. Thus, we directly fuse low-level feature En1-2 with high-level feature En1-4, to propagate the high-level semantic information to the low-level. As different channels of feature in high-level response to different semantics, it is unwise to treat all channels without distinction. To alleviate the interference of the irrelevant semantic information, we use channel-wise attention on En1-4 to assign larger weights to channels which are more important for matting task. The operations in channel-wise attention are with low time complexity. Therefore, it demands negligible computational cost. \\ 
\vspace{-0.7cm}
\subsection{Matting Refine Network}
MRN aims at regressing the alpha values of pixels in unknown regions. The MRN takes the concatenation of 3-channel images and the 3-channel segmentation results from SN as 6-channel input. For easy remembering, we named unknown regions as $U$. We design an efficient non-local attention module which can help better regress the alpha values with less computational cost. \\
\textbf{Efficient Non-Local Attention.} Modeling the relevance of different pixels can help regress the accurate alpha value. Because it can help the network effectively group pixels with similar texture, then pay more attention to these pixels. Inspired by~\cite{DBLP:conf/cvpr/0004GGH18}, we use non-local attention to achieve this goal. Common non-local attention can be defined as:
\begin{equation}
    A=Sim(Q,K)V,
\end{equation}
where $Q,K$ and $V$ mean $query$, $key$ and $value$ feature, and $Sim$ is the similarity between query and key feature. 

According to traditional methods, pixels with almost identical texture should have similar opacity. Hence, pixels that share similar texture information should have similar opacity features. However, as the network layers deepen, the feature representations of each pixel will be blurred, which will affect the regression of alpha value. Therefore, we use detailed image features to guide the information flow on Encoder/Decoder features, which improves the regression accuracy of alpha values. In our proposed work, the $query$ feature is the feature from Encoder/Decoder, named as $X_Q$. The $key/value$ feature is the feature extracted from the input image, named as $X_K$ and $X_V$. 

\begin{figure}
    \centering
    \includegraphics[width=\textwidth]{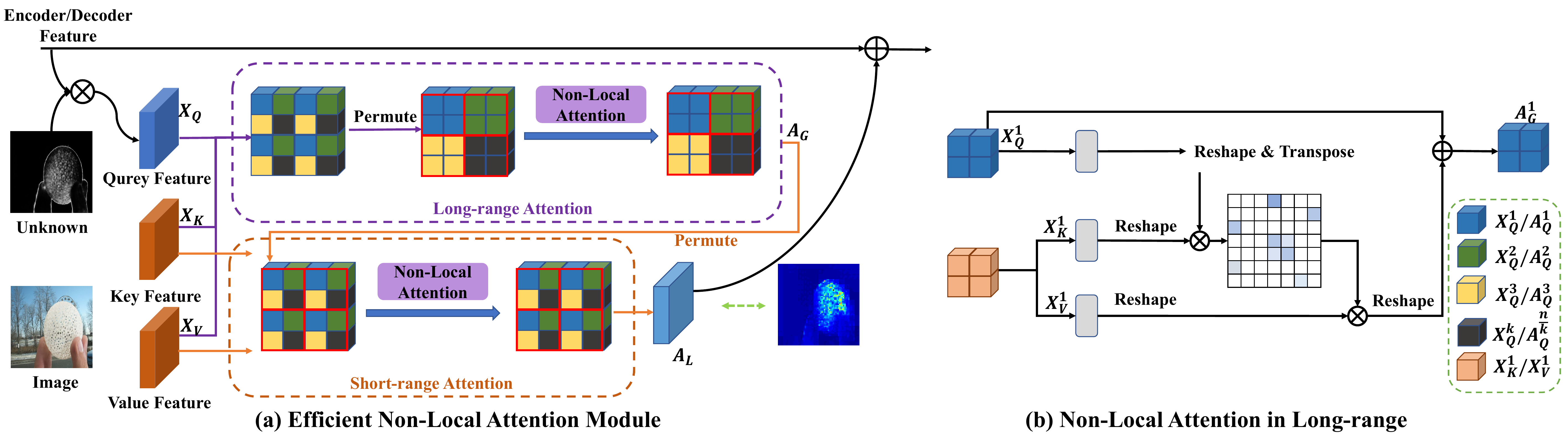}
    \caption{The illustration of the proposed ENA. The query feature is obtained from the multiplication of encoder/decoder features and unknown region. Image is passed through OCBlocks to obtain the key and value feature. Our proposed ENA is a cross-level non-local attention, but for the sake of simplicity, we have only shown the long-range and short-range attention operations on query feature $X_Q$. In this figure, we set $k=4$ and $n=16$.}
    \label{fig:att}
    \vspace{-0.3cm}
\end{figure}

Because MRN only cares about regressing the alpha values in unknown region, we multiply $X_Q$ with $U$ to make $query$ feature more focused on the unknown region. It can be formulated as:
\begin{equation}
    X_Q = X_Q * U.
\end{equation}

Directly using conventional non-local attention in our work is time-consuming, which is not in line with the design principle of light-weighted models. Moreover, motivated by~\cite{DBLP:conf/cvpr/ChenLT12}, modeling the relevance of different pixels in long-range is not as important as modeling the relevance of pixels in short-range, and sometimes will lead to some noise because of irrelevant matches. Thus, our proposed ENA is the combination of long-range attention with sampling strategy and short-range attention. The main point of the long-range attention is to apply the non-local attention to the subsets of positions that have long spatial interval distances, which can not only help reduce the probability of irrelevant matches between long-range pixels, but also save computational cost. The details of long-range attention can be seen in Fig.\ref{fig:att}(a). The input feature $X\in\{X_Q,X_K,X_V\}$ are first divided into $k$ partitions, then sampling these pixels in different partitions at intervals $\sqrt{k}$ into new subsets. In our work, we use permute operation on behalf of sampling. So each subset contains $\frac{n}{k}$ pixels, 
where $n$ represents the size of input features. Thus the new subsets are defined as: $X_L=[X^1,X^2,\cdots,X^k]$, where $X^k\in\{X_Q^k,X_K^k,X_V^k\}$, and the shape of $X^k$ is $\mathbb{R}^{\frac{n}{k}\times C}$. $C$ means the channel numbers. Then we apply the non-local attention on $X^k$ independently:
\begin{equation}
    A_G^k=Sim(X_Q^k, X_K^k)X_V^k,
\end{equation}
which is shown in Fig.\ref{fig:att}(b). Finally, we merge all the $A_G^k$ to get the output $A_G$. For short-range attention, we permute the $A_G$ to group the originally nearby positions together, which is the $query$ of short-range attention. So the input of short-range attention is $Z\in\{A_G,X_K,X_V\}$. Similarly, We divide the into $\frac{n}{k}$ partitions and each partition contains $k$ neighboring positions, defined as: $Z_S=[Z^1,Z^2,\cdots,Z^\frac{n}{k}]$, where $Z^\frac{n}{k}\in\{A_G^\frac{n}{k},X_K^\frac{n}{k},X_V^\frac{n}{k}\}$, and the shape of $Z^\frac{n}{k}$ is $\mathbb{R}^{k\times C}$. Then we apply the non-local attention on each $Z^\frac{n}{k}$ independently. Accordingly, we can get $A_L^\frac{n}{k}$ and merge all of them to get the output $A_L$. Finally, we directly fuse the $A_L$ with Encoder/Decoder features to get the refined features. In our proposed model, we set $k=16$. \textbf{We analyze the computation/memory cost of the common non-local attention mechanism and our ENA in supplementary materials Section 3.} \\
\textbf{MRN Output.} Together with unknown maps, MRN will output the refined alpha values $\alpha_{r}$ in unknown regions. The final alpha matte result $\alpha$ can be written as: 
\begin{equation}
    \alpha=F+\alpha_{r}.
\end{equation}

\vspace{-0.3cm}
\subsection{Loss Function}
To supervise our proposed network, the loss function contains two parts: $L_{SN}$ and $L_{MRN}$. 
Let $\alpha^{gt}$ be the groundtruth of final prediction $\alpha$.
\\
\quad The prediction of SN is 3-channel classification logits $T$. Following~\cite{DBLP:conf/iccv/CaiZFHLLLWS19}, we generate its groundtruth $T^{gt}$ according to $\alpha^{gt}$, which is defined as:
\begin{equation}
	T^{gt}(x,y) = 
	\begin{cases}
	0, \  &{\alpha^{gt}(x,y) = 0} \\
	1, \  &{ \alpha^{gt}(x,y) = 1} \\
	2, \  &{ 0 < \alpha^{gt}(x,y) < 1}
  \end{cases}
\end{equation}
where $(x,y)$ stands for each pixel location on the image. The value $\{0, 1, 2\}$ means the pixel is in background, foreground or unknown region. We use Softmax cross-entropy loss to supervise SN, which is defined as:
\begin{equation}
\begin{aligned}
L_{SN} = \sum_l^{l=3} CE(T_l,T_{gt}).
\end{aligned}
\end{equation}
As show in Fig.\ref{fig:pipeline}, we use multi-level supervision as an auxiliary loss to facilitate suﬃcient training. $T_l$ means the prediction result in $l-$level. 
\\
\quad To supervise MRN, following \cite{DBLP:conf/cvpr/ZhangGFRHBX19}, we adopt the weighted $L_1$ loss and gradient loss $L_g$. The weighted $L_1$ loss is defined as:
\begin{equation}
    L_1 = \sum_p w_p \cdot |\alpha_p - \alpha_p^{gt}|,
\end{equation}
where $\alpha_p$ means predicted alpha value at pixel $p$. The weight $w_p$ is set to 1 when $0<\alpha_p<1$ and set to 0.1 when $\alpha_p=1$ or $\alpha_p=0$. This operation can make the network pay attention to unknown region. The gradient loss $L_g$ is used to remove the over-blurred alpha matte, which is defined as:
\begin{equation}
    L_g =  | \nabla_x(\alpha) - \nabla_x(\alpha^{gt})| + | \nabla_y(\alpha) - \nabla_y(\alpha^{gt})|.
\end{equation}
\\
\quad Thus $L_{MRN}$ can be written as: $L_{MRN} = L_1 + L_g.$
Note that all parts of our proposed network are trained jointly, so the overall loss function is given as: 
\begin{equation}
    L = L_{SN} + L_{MRN}.
\end{equation}
\vspace{-0.6cm}
\section{Experiments}
\subsection{Dataset and Evaluation Metrics}

\textbf{Dataset}. We evaluate our proposed network on the public Adobe Composition-1k \cite{DBLP:conf/cvpr/XuPCH17}. The training set consists of 431 foreground objects with the corresponding ground truth alpha mattes. Each foreground image is combined with 100 background images from MS COCO dataset \cite{DBLP:conf/eccv/LinMBHPRDZ14} to composite the input images. For the test set, the Composition-1k contains 50 foreground images as well as the corresponding alpha mattes, and 1000 background images from PASCAL VOC2012 dataset \cite{DBLP:journals/ijcv/EveringhamGWWZ10}. The training and test sets were synthesized through the algorithm provided by \cite{DBLP:conf/cvpr/XuPCH17}.

For the data augmentation, following \cite{DBLP:conf/aaai/LiL20a}, the foreground object and the alpha image will be resized to 640 $\times$ 640 images with a probability of 0.25 to make the network see the whole foreground image instead of a cropped snippet. Then, a random affine transformation is used. Subsequently, we randomly crop one 512 $\times$ 512 patch from each foreground. All of the patches are centered on an unknown region according to the alpha images. The foreground images are then converted to HSV space, and different jitters are imposed to the hue, saturation, and value. During testing, the resolution of input image is $512\times512$. \\
\textbf{Evaluation metrics}. There are four metrics used in the evaluations: SAD (sum of absolution difference), MSE (mean square error), gradient, and connectivity defined in \cite{DBLP:conf/cvpr/XuPCH17}. The lower values of the metrics, the better the predicted alpha matte is. The details of gradient and connectivity metrics can be found in \cite{DBLP:conf/cvpr/RhemannRWGKR09}, they are used to reflect the visual quality of the alpha matte when observed by a human. \\
\textbf{Implementation Details}. We use Pytorch to implement our model. The whole network is trained in an end-to-end manner. For loss optimization, we use the Adam optimizer with $\beta_1=0.5$ and $\beta_2=0.999$. The learning rate is initialized to $4\times10^{-4}$. Warmup and cosine decay \cite{DBLP:conf/iclr/LoshchilovH17,DBLP:conf/cvpr/HeZ0ZXL19} are applied to the learning rate. The network is trained for 70 epochs with a batch size of 12. We set the weight for $L_{SN}$ to 1, for $L_g$ to 0.5, and for $L_1$ to 5. If not specially mentioned, we use this as the default settings. Our model can be trained from scratch without other datasets.

\begin{figure}[t]
    \centering
    \includegraphics[width=\textwidth]{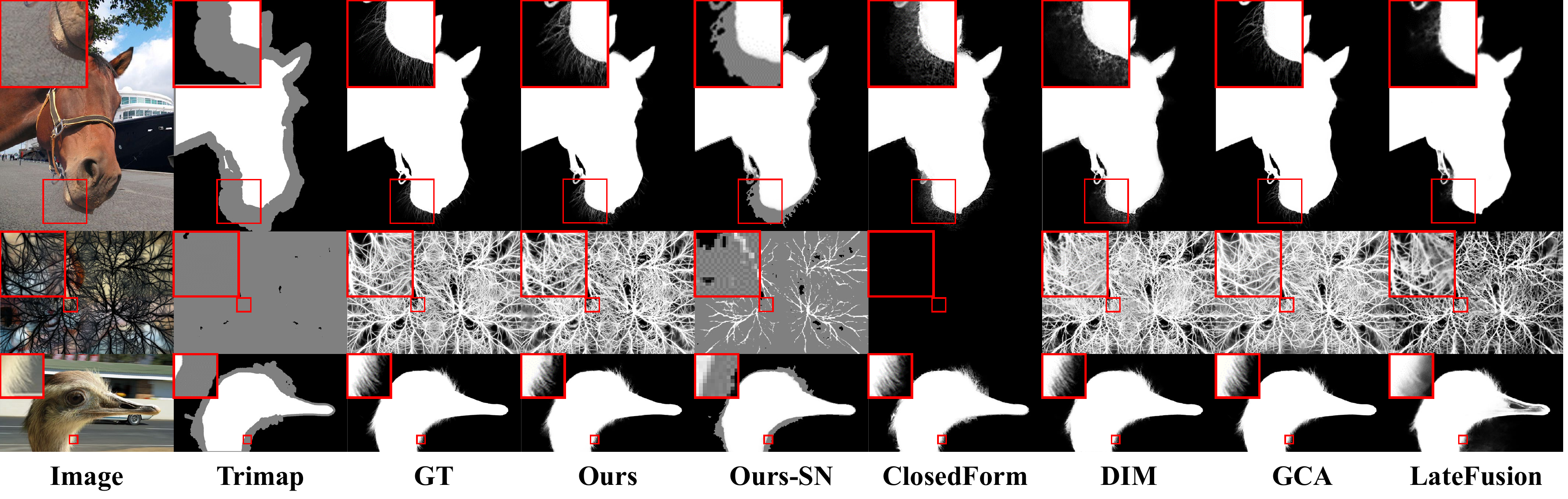}
    \caption{The visual comparisons on the Composition-1k testing dataset. Zoom in for the best view.}
    \label{fig:res}
\end{figure}

\subsection{Evaluation on Composition-1k Test Set}

Here we compare our network with 5 traditional matting methods: Shared Matting \cite{DBLP:journals/cgf/GastalO10}, Learning Based \cite{DBLP:conf/iccv/ZhengK09}, Global Matting \cite{DBLP:conf/cvpr/HeRRTS11}, ClosedForm \cite{DBLP:conf/cvpr/LevinLW06}, KNN Matting \cite{DBLP:conf/cvpr/ChenLT12}, and 7 deep learning based methods: DCNN \cite{DBLP:conf/eccv/ChoTK16}, DIM \cite{DBLP:conf/cvpr/XuPCH17}, AlphaGAN \cite{DBLP:conf/bmvc/LutzAS18}, 
IndexNet \cite{DBLP:conf/iccv/0003DSX19}, Late Fusion \cite{DBLP:conf/cvpr/ZhangGFRHBX19}, HAttMatting \cite{DBLP:conf/cvpr/QiaoLYZXZW20}, GCA \cite{DBLP:conf/aaai/LiL20a}. LateFusion, HAttMatting and our proposed network are trimap-free matting methods. For other methods, we feed RGB images and trimaps produced by 25 pixels random dilation refer to \cite{DBLP:conf/cvpr/XuPCH17}. We compute the metrics at $800\times800$ for all methods. The testing settings in our work are following two trimap-free methods ~\cite{DBLP:conf/cvpr/ZhangGFRHBX19,DBLP:conf/cvpr/QiaoLYZXZW20}. The qualitative results are illustrated in Fig.\ref{fig:res} and quantitative comparisons are reported in Tab.\ref{tab:total}. Because we cannot obtain the executable code or results of HAttMatting, we do not show their qualitative results and directly cite quantitative results from their original paper. 

\begin{wraptable}{r}{0.55\linewidth}
\centering
\caption{Quantitative and complexity comparison on the Composition-1k testing dataset. 'DIM-Trimap-less' denotes the results of the DIM method without trimap as input during training. The underline indicates the best of the trimap-based methods. The bolded numbers represent the best performace in trimap-free methods.}
\resizebox{\linewidth}{25mm}{
\begin{tabular}{||lcccccc||}
\hline
\multicolumn{1}{||c}{\multirow{2}{*}{Model}} & \multicolumn{2}{c}{Complexity} & \multicolumn{4}{c||}{Composition-1k} \\
\multicolumn{1}{||c}{} & \#PARM. & FLOPs & SAD & MSE & Grad & Conn \\ \hline  \hline
Shared Matting & - & - & 115.20 & 0.031 & 139.88 & 121.35 \\ 
Learning Based & - & - & 100.51 & 0.027 & 107.83 & 104.74 \\
Global Matting & - & - & 121.46 & 0.034 & 125.11 & 133.23 \\ 
Closed-form & - & - & 118.98 & 0.032 & 130.77 & 118.17 \\ 
KNN Matting & - & - & 133.99 & 0.051 & 140.29 & 134.03 \\ 
DCNN & - & - & 122.40 & 0.030 & 129.57 & 121.80 \\ \hline \hline
DIM-Trimap-less & 130.55M & 443.56G & 103.07 & 0.108 & 73.76 & 106.64 \\ 
DIM & 130.55M & 443.56G & 33.64 & 0.008 & 30.23 & 31.92 \\ 
IndexNet & 5.72M & 71.41G & 14.97 & \uline{0.006} & 12.02 & 14.58 \\ 
AlphaGAN & 40.3M & 496.36G & 20.28 & 0.010 & 21.11 & 20.46 \\ 
GCA & 25.16M & 89.41G & \uline{13.54} & \uline{0.006} & \uline{11.39} & \uline{12.96} \\ \hline \hline
Late Fusion & 37.91M & 74.85G & 58.29 & 0.011 & 36.58 & 59.63 \\ 
HAttMatting & 59.2M & 159.7G & 44.01 & 0.007 & 29.26 & 46.41 \\
Ours & 344.82K & 0.93G & \textbf{19.59} & \textbf{0.006} & \textbf{18.70} & \textbf{20.34} \\ \hline
\end{tabular}}
\label{tab:total}
\end{wraptable}

As can be seen in Tab.\ref{tab:total}, our proposed method outperforms all the traditional image matting methods by a large margin. Compared to trimap-based deep-learning methods, our method is slightly inferior to GCA and IndexNet, and is better than DIM and AlphaGAN. However, trimaps are strongly required during their training and inference phase, which restricts their effectiveness in practical applications. Compared to trimap-free methods LateFusion and HAttMatting, our proposed method can already outperform these two methods. Moreover, our method achieves SOTA performance with fewer parameters ($0.3\% \thicksim 7.8\%$ of large models) and FLOPs ($0.2\% \thicksim 1\%$ of large models) than all of the above methods which will upgrade high practicality. 

In Fig.\ref{fig:res}, compared to LateFusion, our method is able to find a more complete foreground of the matte objects since our network can exploit multi-scale features to better capture the semantics. Because of our proposed ENA, our method can regress more accurate alpha values in the unknown region.

\vspace{-0.3cm}
\subsection{Ablation Study}
\vspace{-0.3cm}

\begin{wraptable}{r}{0.55\linewidth}
\centering
\caption{Ablation study on CFM and ENA. We also replaced our backbone with the existing lightweight backbone.}
\resizebox{\linewidth}{11mm}{
\begin{tabular}{||c|cc|cccccc||}
\hline
\multirow{2}{*}{Backbone} & \multicolumn{2}{c|}{Configurations} & \multicolumn{2}{c}{Model} & \multicolumn{4}{c||}{Metrics} \\
 & CFM & ENA & \# Param. & FLOPs & SAD & MSE & Grad & Conn \\ \hline \hline
\multirow{4}{*}{Ours} & $\times$ & $\times$ & 285.70K & 0.63G & 56.86 & 0.019 & 36.81 & 48.16 \\
 & $\times$ & $\checkmark$ & 342.70K & 0.93G & 34.33 & 0.021 & 28.24 & 32.98 \\
 & $\checkmark$ & $\times$ & 287.83K & 0.63G & 28.90 & 0.013 & 24.27 & 26.35 \\
 & $\checkmark$ & $\checkmark$ & 344.93K & 0.93G & \textbf{19.59} & \textbf{0.006} & \textbf{18.70} & \textbf{20.34} \\ \hline \hline
MobileNetV2 & $\checkmark$ & $\checkmark$ & 37.76M & 27.02G & 47.21 & 0.014 & 30.28 & 36.46 \\
ShuffleNetV2 & $\checkmark$ & $\checkmark$ & 30.22M & 20.52G & 48.72 & 0.015 & 33.84 & 38.21 \\ \hline
\end{tabular}}
\label{tab:ab1}
\end{wraptable}

The core idea of our method contains three parts: light-weighted backbone, cross-level fusion module (CFM), and Efficient Non-Local Attention (ENA). To validate the effectiveness of our proposed light-weighted backbone, we replace it with other light-weighted backbones, including MobileNet and ShuffleNet, and results are shown in Tab.\ref{tab:ab1}. Compared to these two light-weighted backbones, our proposed backbone can achieve better performance with fewer parameters and FLOPs. Because our backbone can achieve multi-scale feature representations for better capturing sufficient semantics. Moreover, only using CFM or ENA can already heavily improve the performance. Better performance has been achieved through the combination of these two architectures. 

\begin{figure}[!t]
    \centering
    \includegraphics[width=\textwidth]{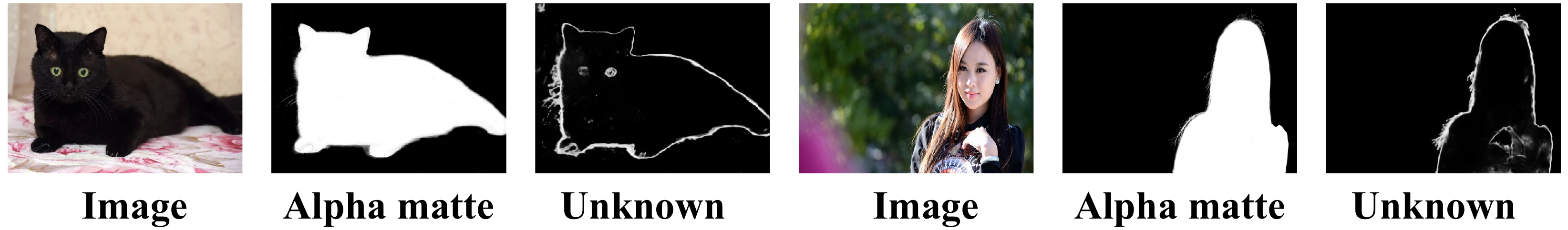}
    \caption{The results on real world images and the unknown region predicted by our method.}
    \label{fig:real}
    \vspace{-0.3cm}
\end{figure}

To further verify the validity and reasonableness of our designed efficient non-local attention module. We trained our model with different types of attention: common non-local attention and efficient non-local attention model. As shown in Tab.\ref{tab:ab2}, compared to common-local attention, our proposed ENA is not only more efficient but also more usable for matting tasks. Using only short-range attention (S) or long-range attention (L) can improve model performance because they can model the relevance between different pixels, which can help regress alpha values. As can be seen in Tab.\ref{tab:ab2}, different sampling intervals $\sqrt{k}$ lead to different results. When $\sqrt{k}=1,2$, ENA cannot fully exploit its convenience and effectiveness, because the sampling is too intensive. And the long-range attention will lead to some artifacts because of irrelevant matches. Experimentally, we set $\sqrt{k}$ to 4 in all experiments. For more comprehensive analyses of our proposed method, please refer to the supplementary materials.
Fig.\ref{fig:real} shows the matting results for internet images for further testing of our method. 

\begin{table}[t]
\centering
\caption{Ablation study for different modalities using attention. $S$ and $L$ means short- and long-range attentions respectively.}
\resizebox{\linewidth}{21mm}{
\begin{tabular}{||cc|cccccc||}
\hline
\multicolumn{2}{||c|}{\multirow{2}{*}{Configurations}} & \multicolumn{2}{c}{Model} & \multicolumn{4}{c||}{Metrics} \\
\multicolumn{2}{||c|}{} & \# Param. & FLOPs & SAD & MSE & Grad & Conn \\ \hline \hline
\multicolumn{2}{||c|}{Common Non-Local} & 421.69K & 1.33G & 25.13 & 0.010 & 20.89 & 22.67 \\ \hline \hline
\multirow{6}{*}{Efficient Non-local(Ours)} & \multicolumn{1}{l|}{S} & 325.15K & 0.82G & 25.98 & 0.011 & 21.93 & 23.02 \\
 & \multicolumn{1}{l|}{L($\sqrt k$=4)} & 292.19K & 0.72G & 27.01 & 0.013 & 23.94 & 24.31 \\
 & S+L($\sqrt k$=1) & 424.11K & 1.34G & 25.01 & 0.010 & 20.31 & 22.96 \\
 & S+L($\sqrt k$=2) & 371.43K & 1.02G & 23.27 & 0.009 & 19.37 & 21.61 \\
 & S+L($\sqrt k$=4) & 344.83K & 0.93G & \textbf{19.59} & \textbf{0.006} & \textbf{18.70} & \textbf{20.34} \\
 & S+L($\sqrt k$=8) & 340.27K & 0.91G & 19.90 & 0.008 & 19.05 & 21.35 \\ \hline
\end{tabular}}
\label{tab:ab2}
\vspace{-0.3cm}
\end{table}
\vspace{-0.5cm}
\section{Conclusion}
\vspace{-0.3cm}
In this paper, we design a two-stage trimap-free light-weighted natural image matting model, which makes one of the earliest effort on this area. SN is designed to capture sufﬁcient semantics and classify the pixels into unknown, foreground and background regions. MRN aims at capturing detailed texture information and regressing accurate alpha values. With proposed CFM, SN can efficiently utilize multi-scale features with less computational cost. ENA in MRN can efficiently model the relevance between different pixels and help regress high-quality alpha values. Experiment results on Composition-1k testing set demonstrate that our method outperforms state-of-the-art (SOTA) approaches in natural image matting with ~1\% parameters (344k) of large models. We believe our proposed backbone, CFM and ENA can contribute to other light-weighted computer vision tasks in the future.

\bibliography{egbib}
\end{document}


\maketitle

\section{Introduction}

This supplemental material contains 6 parts:

\begin{itemize}
    \item Section 2 gives more details about our proposed lightweight backbone.
    \item Section 3 gives more analysis of the proposed ENA.
\end{itemize}

We hope this supplemental material can help you get a better understanding of our work.

\section{More Details of Our Network}

\subsection{Details of Network Architecture}

\begin{figure}[ht!]
    \centering
    \includegraphics[width=\textwidth]{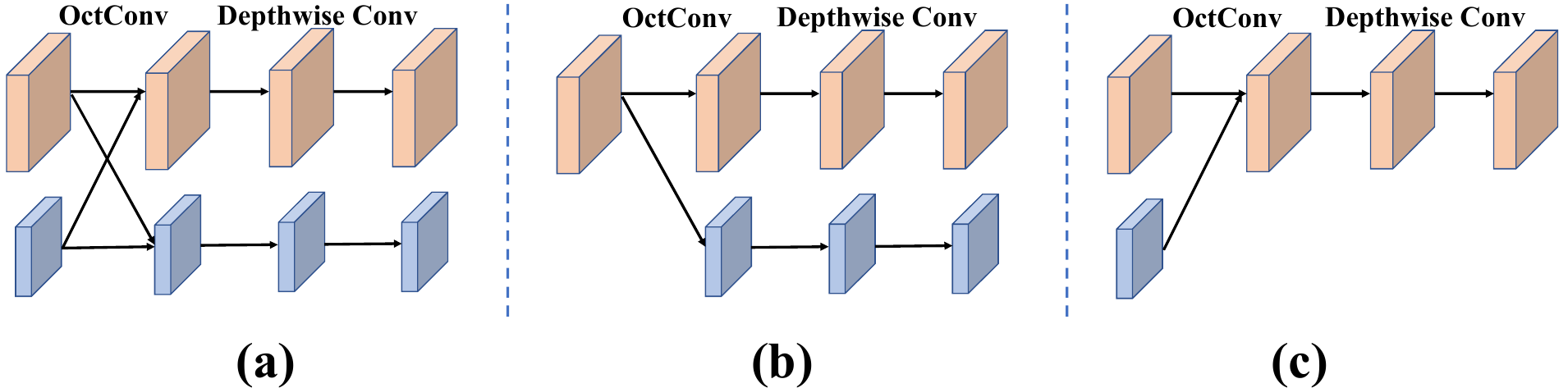}
    \caption{Different types of OCBlock.}
    \label{fig:net}
\end{figure}

When OCBolck has only one scale input, OCBlock will be built like Fig.\ref{fig:net}(b) to obtain multi-scale output. When OCBlock has only one scale output, it will be built like Fig. \ref{fig:net}(c) to fuse the multi-scale input. The general OCBlock is shown in Fig. \ref{fig:net}(a).

Specifically, in stage1 of the proposed network, we use another OCBlock like Fig.\ref{fig:net}(b) as En1-0 to decompose the input image into features at two resolutions (256 and 512). So the dimension of the input of our backbone is $[256, 512]$. The dimension output of each level is $[256, 512], [256], [128], [64]$ respectively. We only use the $512$ features from En1-1 as described in main text. Thus, the first OCBlock in En1-3 and En1-4 uses Fig. \ref{fig:net}(b) structure. The last OCBlock in En1-2, En1-3 and En1-4 uses Fig. \ref{fig:net}(c) structure. MRN uses a similar structure.

\section{More analysis of ENA}

\subsection{The Efficiency of ENA}


Given an input feature map of size $H\times W\times C$, we analyze the computation cost of both the common non-local attention mechanism in the whole image and our proposed ENA.

For common non-local attention mechanism, three $1\times 1$ convolution layers acting on $Q,K,V$ to change them from $\mathbb{R}^{C\times H \times W}$ to $\mathbb{R}^{\frac{C}{2}\times H \times W}$ and one $1\times 1$ convolution layers acting on the output feature to change it from $\mathbb{R}^{\frac{C}{2}\times H \times W}$ to $\mathbb{R}^{C\times H \times W}$. The complexity of this part is ${\cal{O}}(2C^2HW)$. And, the complexity of multiplying between features is ${\cal{O}}(\frac{3}{2}C(HW)^2)$. So, the complexity of self-attention mechanism is
\begin{equation}
    C_1 = {\cal{O}}(2C^2HW+\frac{3}{2}C(HW)^2).
\end{equation}

Considering our approach, we divide the height and width dimension to $\sqrt{k}$ groups in calculating long-range relations, and $\frac{H}{\sqrt{k}}$ and $\frac{W}{\sqrt{k}}$ groups in calculating short-range relations. So the complexity of our approach is
\begin{equation}
    C_2 = {{\cal{O}}(4C^2HW+\frac{3}{2}C(HW)^2(\frac{1}{k}+\frac{k}{HW}))}.
\end{equation}
Since the channel number $C$ is a smaller value in our approach, it is much smaller than $HW$. The increase in the first term is negligible compared to the decrease in the second term. The complexity of our approach can be minimized to
\begin{equation}
    min(C_2) = {{\cal{O}}(4C^2HW+3C(HW)^\frac{3}{2})}.
\end{equation}
when $k=\sqrt{HW}$. Although $\sqrt{k}$ will take a small value in the implementation (e.g. 4) and will not be able to minimize $C_2$, our approach is efficient enough compared to original self-attention.


For the numerical complexity, we consider the feature dimensions of 80 and $\sqrt{k}=4$. When the input size is $64\times 64$, the computation cost of the self-attention mechanism is 2.06 GFLOPs and ours ENA is 0.11 GFLOPs. When the input size is $128\times 128$, it becomes 32.42 GFLOPs and 0.45 GFLOPs. So, our proposed efficient non-local attention module not only guarantees a small amount of computation cost when the input size is small but also guarantees a slow growth when the input size increases.

\subsection{The location choice of ENA in our network}

CFM wants to avoid the gradual dilution of the high-level semantic information
during decoding and propagate it to the low-level. Multiple CFMs will lead to duplication of operations and produce redundant computations, so we only build CFM between En1-2 corresponding to the last decoder and En1-4. For the ENA module, our experiments show that putting it at En/De2-2 does not improve performance while increasing complexity.